\RequirePackage{fix-cm}
\documentclass{svjour3}                     % onecolumn (standard format)
\smartqed  % flush right qed marks, e.g. at end of proof
\usepackage{graphicx}
\usepackage[colorlinks,citecolor=blue,urlcolor=blue,filecolor=blue,backref=page]{hyperref}
\usepackage{amsmath}
\usepackage[utf8]{inputenc} % allow utf-8 input
\usepackage[T1]{fontenc}    % use 8-bit T1 fonts
\usepackage{amssymb} 
\usepackage{wrapfig}
\usepackage{enumitem}
\usepackage{booktabs,etoolbox}
\usepackage{xcolor}
\usepackage{amsbsy}
\usepackage{mathtools}
\usepackage{subcaption}
\usepackage[ruled,linesnumbered]{algorithm2e}
\usepackage{url}            % simple URL typesetting
\usepackage{booktabs}       % professional-quality tables
\usepackage{amsfonts}       % blackboard math symbols
\usepackage{nicefrac}       % compact symbols for 1/2, etc.
\usepackage{natbib}
\usepackage{microtype}      % microtypography
\usepackage{epsfig}
\usepackage{epsf}
\usepackage{manfnt}
\usepackage{color}

\newcommand{\refp}[1]{(\ref{#1})}
\makeatletter
\newcommand{\oset}[3][0ex]{%
  \mathrel{\mathop{#3}\limits^{
    \vbox to#1{\kern-2\ex@
    \hbox{$\scriptstyle#2$}\vss}}}}
\makeatother

\newcommand{\ie}{\emph{i.e.}}
\newcommand{\eg}{\emph{e.g.}}

\newcounter{xxx}
\begin{document}

\title{Shape Modeling with Spline Partitions%\thanks{Grants or other notes
}
%\subtitle{SMSP}


\author{Shufei Ge         \and
        Shijia Wang \and Lloyd Elliott%etc.
}


\institute{S. Ge \at
              Institute of Mathematical Sciences, ShanghaiTech University, China\\
               \email{geshf@shanghaitech.edu.cn}           %  \\
           \and
           S. Wang \at
              School of Statistics and Data Science, LPMC and KLMDASR, Nankai University, China\\
               \email{shijia\_wang@nankai.edu.cn}               \and
           L. Elliott \at
              Department of Statistics and Actuarial Science, Simon Fraser University, Canada\\
               \email{lloyd\_elliott@sfu.ca}\\
}

\date{Received: date / Accepted: date}

\maketitle

\begin{abstract}
Shape modelling (with methods that output shapes) is a new  and important task in Bayesian nonparametrics and bioinformatics. In this work, we focus on Bayesian nonparametric methods for capturing shapes by partitioning a space using curves.
In related work, the classical Mondrian process is used to partition spaces recursively with axis-aligned cuts, and is widely applied in multi-dimensional and relational data. The Mondrian process outputs hyper-rectangles. Recently, the random tessellation process was introduced as a generalization of the Mondrian process, partitioning a domain with non-axis aligned cuts in an arbitrary dimensional space, and outputting polytopes. Motivated by these processes, in this work, we propose a novel parallelized Bayesian nonparametric approach to partition a domain with curves, enabling complex data-shapes to be acquired. We apply our method to HIV-1-infected human macrophage image dataset, and also simulated datasets sets to illustrate our approach. We compare to support vector machines, random forests and state-of-the-art computer vision methods such as simple linear iterative clustering super pixel image segmentation. 
We develop an R package that is available at \url{https://github.com/ShufeiGe/Shape-Modeling-with-Spline-Partitions}.
%\subclass{MSC code1 \and MSC code2 \and more}
\end{abstract}

\section{Introduction}
Shapes (the boundaries of objects) within images are widely studied in biological science. This application domain includes for example classifying cells collected in a blood sample from individuals, and characterizing brain diseases of patients by analyzing shapes within brain images. One basic issue that arises in estimating shapes is the choice of the representation of the shape. Curves \citep{bu2014computational, muller1997surface} are commonly used to provide flexible representations of a wide range of objects (\eg~tumors, cells). In related work, shape models have been explored using smooth stochastic processes \citep{kurtek2012statistical}.
 Bayesian hierarchical models and Gaussian processes have also been used  \citep{gu2012bayesian,lin2019extrinsic}. Bayesian nonparametrics has identified shape modelling as an important area of application, supplementing traditional classification, regression and clustering applications~\citep{hannah2013multivariate,bhattacharya2010nonparametric}. We consider modeling shapes within image with curves, in the framework of space partitioning methods.

The Mondrian process~\citep{roy2008mondrian} partitions multidimensional space into hyper-rectangles with recursive axis-aligned hyperplane cutting.  In the Mondrian process, hyper-rectangles are assigned to  lifetimes (\emph{i.e.}, costs) sampled from an exponential distribution, with rate given by the perimeter of the hypercube.  
 In each iteration, the hypercube with the smallest lifetime is divided into two new hyper-rectangles by a randomly sampled hyperplane. New lifetimes for all existing hyper-rectangles are then sampled.  This process repeats until the total cost exceeds a prespecified budget.  The Mondrian process is used as a nonparametric prior distribution in Bayesian models of relational data, and provides a Bayesian view of decision trees~\citep{roy2008mondrian,kemp2006learning}.  Developments in the Mondrian process for space partitioning achieve high predictive accuracy, with improved  efficiency. These methods include online methods~\citep{lakshminarayanan2014mondrian}, and particle Gibbs  for Mondrian process additive regression trees~\citep{lakshminarayanan2015particle}.  However, the axis-aligned hyperplanes of the decision boundaries of the Mondrian process restrict its flexibility, which could lead to failure in capturing inter-dimensional dependencies  among features.   Bayesian nonparametric methods have been introduced to generalise the Mondrian process and allow non-axis aligned cuts, such as generating arbitrary oblique cutting lines or hyperplanes \citep{fan2016ostomachion, fan2018binary, fan2018rectangular, bspf, fan2020online,  ge2019random} to partition the target space.  Alternative constructions of non-axis aligned partitioning for multidimensional spaces and non-Bayesian methods  include sparse linear combinations of predictors or canonical correlation as random forest generalisations~\citep{george1987sampling,tomita2015random,rainforth2015canonical}.
  
\begin{figure}[ht]
\centering
  \includegraphics[width=.3\linewidth]{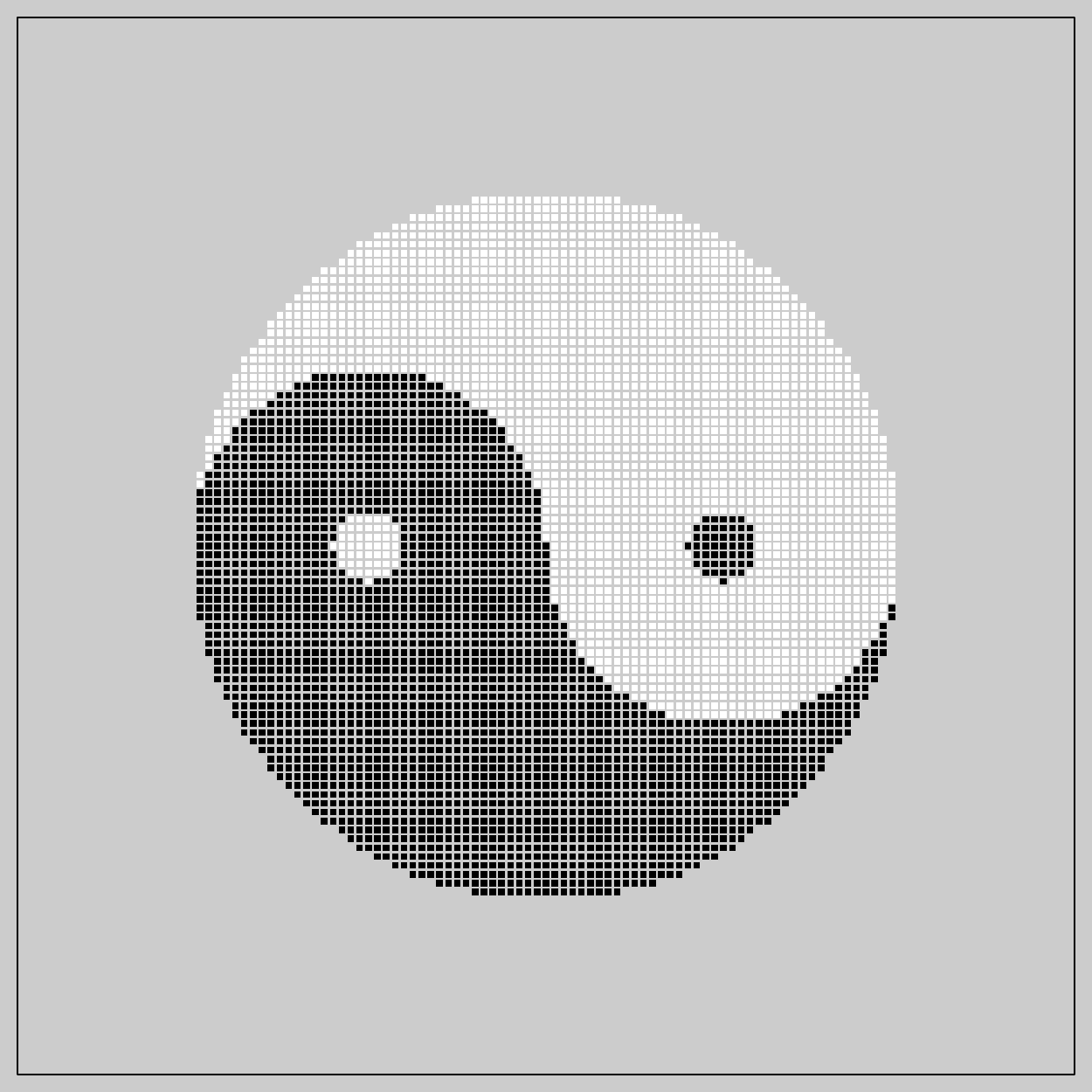}
   
\caption{{\color{black}  A view of the yin-yang full dataset, with black indicating label $1$ and white indicating label $2$. We refer readers to Section \ref{sec:yin-yang} for more details on the simulation setups. 
}}
\label{fig:fig1}
\end{figure}

We propose a \emph{shape modeling with spline partitions} process (SMSP), an extension of the Mondrian process and {random tessellation process} (RTP),  which further modifies the random tessellation process by replacing the non-axis aligned cutting hyperplanes with curved cuts. The SMSP provides a framework for describing Bayesian nonparametric models based on partitioning  two-dimensional Euclidean space with splines. The proposed nonlinear curves enable more complex data-structures to be represented, and allow arbitrary shapes to be approximated and outputted by the model. {\color{black} For example, our simulation studies (Section \ref{sec:yin-yang}) show that SMSP with one cut is able to capture the shapes displayed in Figure \ref{fig:fig1}, and can perform better than one-cut RTP and support vector machines (SVMs) with 
linear, polynomial and sigmoid kernel functions. } As with the random tessellation process, the construction of the SMSP is based on the theory of stable iterated tessellations in stochastic geometry~\citep{nagel2005crack}. The partitions induced by the SMSP prior are described by a sets of  compact subsets. We note that in contrast to the RTP, each compact subset could contain more than one disjoint connected components (\ie, the blocks of the partition need not be connected). 

We implement a sequential Monte Carlo (SMC) algorithm \citep{doucet2000sequential} for SMSP inference, which takes advantage of the hierarchical structure of the generative process of SMSP. The SMC algorithm can be easily parallelized by allocating samples across multiple CPUs (cores) to decrease the computational time.  Our simulation study provides empirical evidence that our SMSP 
 exhibits rotational and translational invariance, however exploring the consistency of SMSP is an area of future work.  We apply our proposed model to simulated yin-yang data and an HIV-1-infected human macrophage dataset to demonstrate its effectiveness.  For the experiment on the macrophage dataset, we consider a task wherein the perimeter of a macrophage cell is estimated. Shapes of biological microstructures may indicate disease etyology. For example, the fractal dimension of vascularization in a cancer tumor is known to be modulated during remission and metastasis~\citep{Gaz1997a}. This motivates perimeter estimation, as estimation of perimeters through a continuum of resolution indicates fractal dimension.

\section{Methods} \label{sec:met}

We characterize shapes by describing their boundaries through a space partitioning approach. Let $(\boldsymbol{v}_1, z_1), \ldots,(\boldsymbol{v}_n , z_n )$  denote $n$ observed data items, where $\boldsymbol{v}_{i}\in\mathbb{R}^2$ are predictors and $z_i \in \{1,\ldots,K\}$ are labels. In our application, we are interested in modeling the shapes within a two-dimensional image. The predictors $\boldsymbol{v}_{i}$ are pixels of image, and the labels $z_i$ are labels of shapes. For example, in our real data application, we are interested in estimating the shapes of HIV-1-infected human macrophage. Pixels falling within the cell are labelled  $1$ and other pixels are labelled $2$.

A Bayesian nonparametric model partitions the feature space by placing a prior distribution on the partition process, and associates model parameters with the blocks of the partition. Inference is then made on the joint posterior of the parameters and the structure of the partition. In this section, we develop the SMSP and place a prior on partitions of $(\boldsymbol{v}_1, z_1), \ldots,(\boldsymbol{v}_n , z_n )$ induced by partitions via splines in a two-dimensional feature space.

\subsection{Shape Modeling with Spline Partitions} \label{sec:rtp}

\begin{figure}[ht!]
 \centering 
 \includegraphics[width=0.99\linewidth]{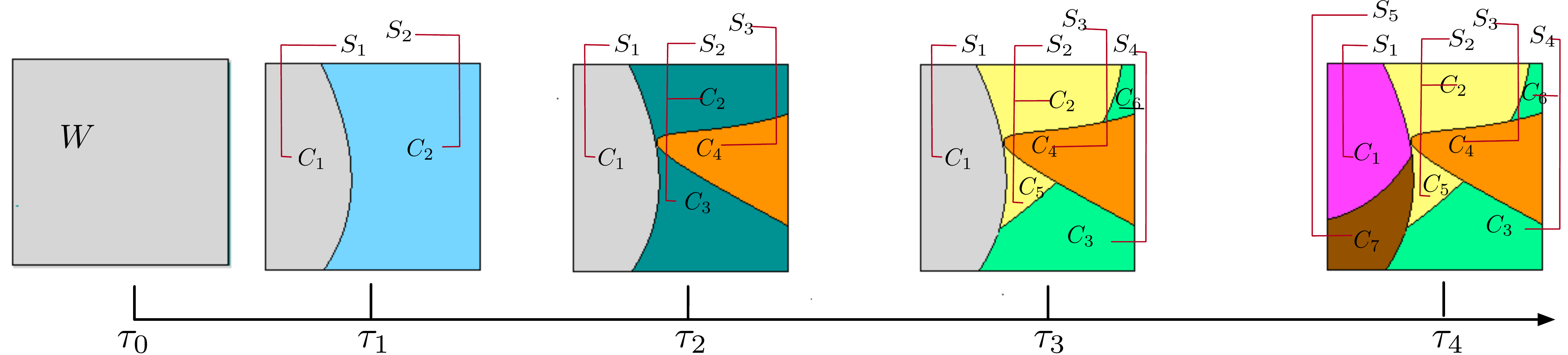}
\caption{ \hspace{-0.15em} An example of a generative process from the \emph{SMSP} prior with domain given by a two dimensional box $(x,y)\!\in\![-1,1]^2$. Colours indicate subsets identity, and are randomly assigned. }
 \label{partt}
\end{figure}

A partition $Y$ in the SMSP of a bounded domain $W \subset \mathbb{R}^2$ is a finite collection of compact subsets, such that the union of the subsets is all of $W$, and the subsets have pairwise disjoint interiors~\citep{stoch}. Note that each subset may contain more than one disjoint connected components, as shown in  Figure~\ref{partt}. Also throughout this work we consider the two-dimensional plane (although, these methods may be generalized to higher dimensions as a point of future work). Denote $\mathbb{C}_0$ as the set of all compact sets with finitely many connected components, $\mathbb{C}_0 = \cup_{j=1}^{J_c} C_j$, $J_c\ge 2$, $\text{interior}(C_i) \cap \text{interior}(C_j) = \varnothing $ for any $i \neq j$. Let $S_i,~ i=1,\cdots,J_s$ denotes each subset in the partition, and each $S_i$ could contain more than one disjoint connected components. We denote partitions of $W$ by $Y\!(W)$,  $Y\!(W) = \cup_{i=1}^{J_s}S_i,~ S_i=\cup_{C_j \in S_i} {C_j}$. For example, in Figure~\ref{partt}, at time $\tau_4$, $Y\!(W) = \cup_{i=1}^{5}S_i,~S_1=C_1,~S_2={C_2} \cup {C_5},~ S_3=C_4,~S_4=C_3 \cup C_6, ~S_5=C_7$.

The SMSP forms a partition. An element of the partition is a subset with non-empty interior formed by the intersection of finitely many closed curves. The SMSP itself $Y_t(W)$ is a partition-valued Markov process defined on $[0, \tau]$, with cutting events specified by splines that intersect the partition's subsets. Here $\tau$ is a prespecified budget~\citep{mjp}. The SMSP $Y_0(W)$ is initialized with a single subset $W$ covering all of the observed predictors: $W$ is the convex hull  $\text{hull}\{\boldsymbol{v}_1,\ldots,\boldsymbol{v}_n\}$. 
The Markov process for the SMSP associates each subset with an exponentially distributed lifetime. The subset is replaced by two new subsets at the end of subset's lifetime $t$. 
At time $t$, we sample a spline that intersects the interior of the old subset to form two new subsets.  We evolve the Markov process until we reach the prespecified budget $\tau$.

As in~\cite{ge2019random}, we associate a measure on splines with the generative process described in \emph{Section}~2.3. If $W'$ is a subset of $W$, let $[W']$ be the set of all splines formed by the generative process described in \emph{Section}~2.3 that intersect $W'$. The generative process thus implicitly defines a measure $\Lambda$ on splines such that the density of the spline found through the generative process is $\Lambda(\cdot)/\Lambda([W'])$. We can sample from this density using rejection sampling. 
{\color{black} We could determine whether a spline intersects an arbitrary subset $W'$ by checking if all points in $W'$ are on one side of the curve or not. If all points in $W'$ falls on the same side of the spline, then the curve does not intersect with $W'$, otherwise, it intersects with the $W'$.} 

 To sample an SMSP, we sample a B\'ezier curve \citep{mortenson1999mathematics} to cut a subset. B\'ezier curves are widely used in computer graphic to draw shapes, due to their flexibility in modeling complex shapes and their simple mathematical formulation. Let $B$ be the set of B\'ezier curves in $R^2$. Every  B\'ezier curve $b$ is uniquely defined by a set of control points 
$\mathbf{P}_0, \ldots, \mathbf{P}_n$, where $n$ is the order of the B\'ezier curve. {\color{black}  The order of a B\'ezier curve equals the number of control points minus one.   Let $\mathbf{B}(t)$ denote the Bézier curve determined by a set of control points $(\mathbf{P}_0,\ldots, \mathbf{P}_{n})$. Then $\mathbf{B}(t) = \sum_{i=0}^{n}\mathbf{P}_{i}B_{n,i}(t)$, $0\le t \le 1$, where $B_{n,i}$ is a Bernstein polynomial, $i=0,\ldots,n$.  Note that the control points are not always on curve, and the B\'ezier curve is always inside the convex hull of control points. } In our numerical experiments,  we let $n$  takes values from $1,2,3$ randomly ($1$ for linear,  $2$ for quadratic, $3$ for cubic). 

 Algorithm \ref{constructionofSMSP} depicts pseudo-code of the generative process for the SMSP. 
 
 \hrulefill\\
\begin{algorithm}[H]
\caption{A generative  process for the \emph{SMSP}.}\label{constructionofSMSP}
  {\bfseries Input:} a) A set $W\!$, b) A measure $\Lambda$ on splines, c) A prespecified budget $\tau$.\\
 {\bfseries Output:} A time-varying SMSP $(Y_t)_{0 \leq t \leq \tau}$.\\
 $\tau_0 \leftarrow 0$, $Y_0 \leftarrow \{W\}$.\\
\While{$\tau_0 \le \tau$} {
 Sample $\tau' \sim \text{Exp}\!\left(\sum_{S \in Y_{\tau_0}} \Lambda([S])\right)$.\\
   Set $Y_t \leftarrow Y_{\tau_0}$ for all $t \in (\tau_0,\min\{\tau,\tau_0+\tau'\}]$,
   Set $\tau_0 \leftarrow \tau_0 + \tau'$.\\
\eIf{$\tau_0 \le \tau$}{
    Sample a subset $S$ from the set $Y_{\tau_0}$ with probability proportional to $\Lambda([S])$.\\
    Sample a B\'ezier curve $b$ from $[S]$ according to the probability measure $\Lambda(\cdot \cap [S])/\Lambda([S])$.\\
    $Y_{\tau_0} \leftarrow \left(Y_{\tau_0}/\{ S \}\right) \cup \{S \cap b^-, S \cap b^+\!\}$.\\
    Here $b^-$ and $b^+$ are the plane partitions on either side of $b$.
    }{
      \Return the Markov process sample $(Y_t)_{0 \leq t \leq \tau}$.
}  
} 
\end{algorithm}
\hrulefill

 \subsection{Coordinate rotation}
\label{sec:cutplane}
\begin{figure}[h!]
     \begin{center}
     \includegraphics[width=1\linewidth]{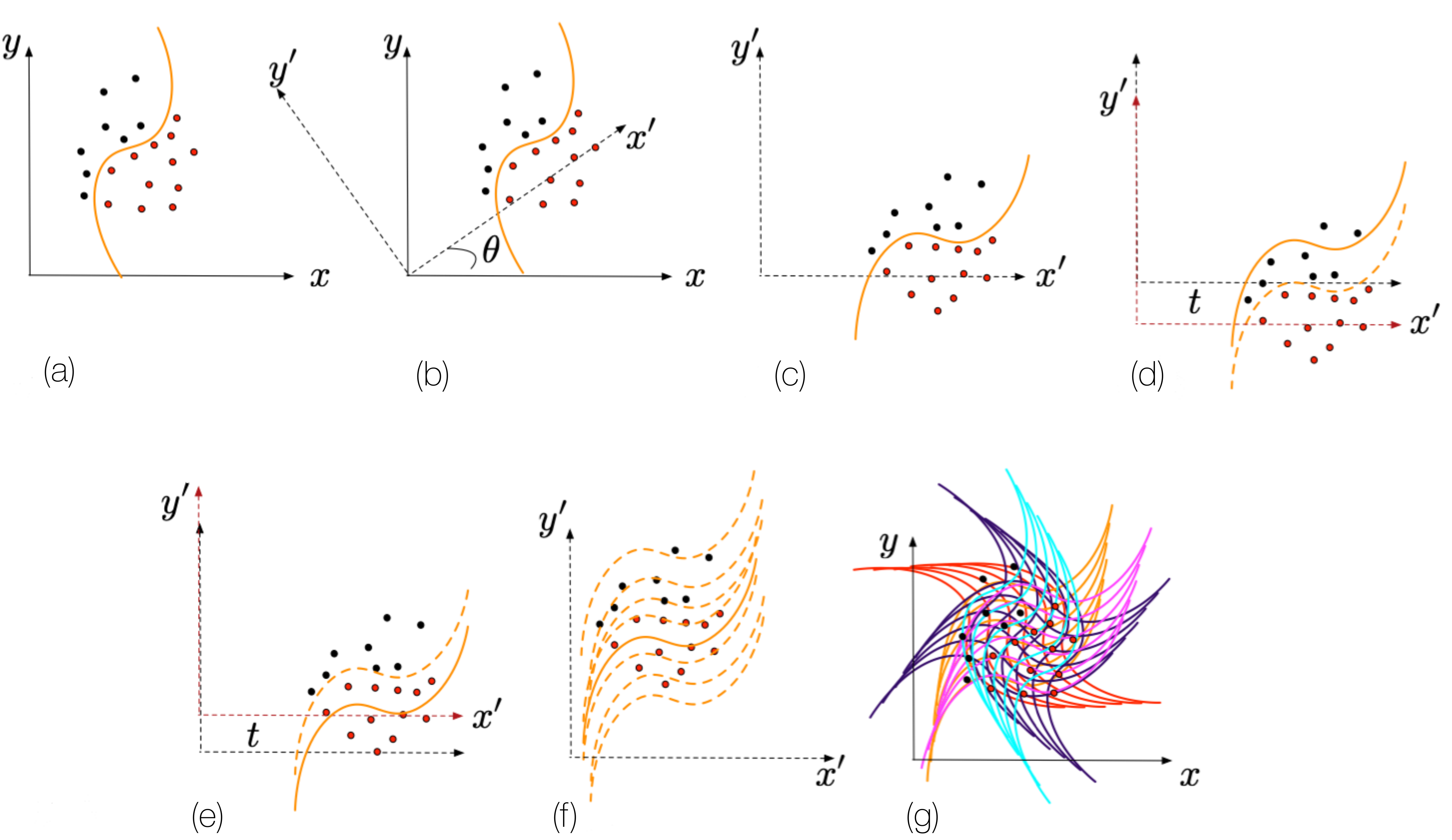}
   \caption{An example of coordinate rotation. (\textit{a}) A rotated B\'ezier curve (with angle $\theta$), which is not  injective  in the original coordinate system. (\textit{b-c}) A new coordinate system and a  B\'ezier curve proposed so as to be injective. The new coordinate is obtained by rotating the original coordinate by angle $\theta$. (\textit{d-e}) Translating a  B\'ezier curve in the new coordinate system. (\textit{f}) A subset of a B\'ezier curve splits the targets from different positioning via moving up and down. (\textit{g}) A subset of a B\'ezier curve splits the target from different angles and positions.} 
 \label{rotate}
    \end{center}
\end{figure}

We discuss the problem of determining which side of a B\'ezier curve a point lies on. This is required in order to partition points in the target, given the B\'ezier curve cuts. After proposing a B\'ezier curve in a reference coordinate system, we may rotate it and translate it along a normal vector in order to produce a proposal for the cut. However, after rotation the curve may no longer be a function of the $x$-axis (\ie, the mapping from $x$ to the rotated and translated spline is not injective), which makes it difficult to identify which side of the curve target points lie on (we cannot just compare their $y$-levels). This is shown in Figure~\ref{rotate} (\emph{a}). In order to let the  B\'ezier curve to split targets after rotation by an arbitrary angle, we rotate the coordinate system rather than the curve. We project the targets to a new coordinate system and the  B\'ezier curve proposal can split the targets by comparing the $y$-levels in the new coordinate system. This rotation is shown in Figure~\ref{rotate} (\emph{b}) - (\emph{c}). 
The translation of the B\'ezier curve is displayed in Figure~\ref{rotate} (\emph{d}) - (\emph{e}).
 Figure~\ref{rotate} (\emph{f}) and (\emph{g}) show subsets of  B\'ezier curves splitting targets according to an arbitrary angle. 

\subsection{Generative process for the cutting spline}

\begin{centering}
\begin{figure}
  \includegraphics[width=1\linewidth]{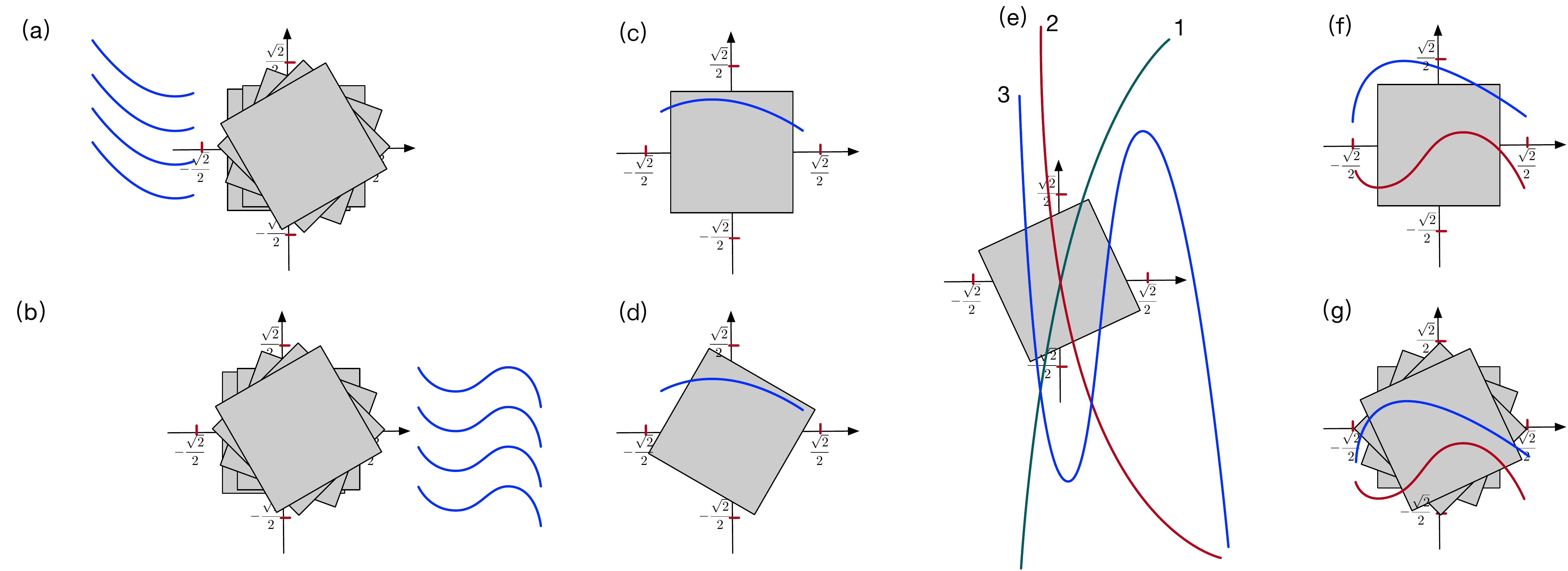}
  \caption{ An illustration for the selection of $a$, $b$, $c$, $d$, the gray cube represents the target to be split and the curved lines are the B\'ezier curves. Panels (\textit{a, b}): If parameters $a, b$ are two small or two large, no matter how we lift the curve up and down, the proposed curve will not intersect with the target; Panels (\textit{c, d}): if $a$ is not small enough or $b$ is not large enough, after rotation, it is likely to result in an invalid intersect. For example as indicated in (\textit{d}), one of the ends of the curve lies in the target, the curve will fail to split the target;  Panel (\textit{e}): if $c$ is too small, or $d$ is too large, the segment of the curves that intersects with target will approximately be straight, it will lose the flexibility of the curve; Panels (\textit{f, g}): with proper $a, b, c, d$, after any arbitrary rotation, by lifting the curve up and down, the proposed curve will split the target successfully once it intersects with the target without losing the flexibility. }
  \label{fig: abcd}
\end{figure}
\end{centering}

 Now we turn to the description of a generative process for the cutting spline. Without loss of generality, we can assume the shape being cut is a square. Our cutting procedure should define a cut given by a single random spline, and incorporate the coordinate rotation scheme. The generative process is as follows:
\begin{enumerate}
\item  Project the shape (i.e., target) to a new coordinate system by rotating the coordinate with angle $\theta$ as in Figure~\ref{rotate} (\emph{a}) - (\emph{c}), and moving it along a normal vector by $t$ as in Figure~\ref{rotate} (\emph{e}) - (\emph{f}). Denote the coordinates before projection by $(x,y)$, and after projection by $(x',y')$.  Without loss of generality, we assume $(x,y)\in [-\frac{1}{2},\frac{1}{2}]\times [-\frac{1}{2},\frac{1}{2}]$. The  rotation proposed above indicates the following transformations:
\begin{align}
\begin{bmatrix} 
x'  \\
y' \\
\end{bmatrix}
=
\begin{bmatrix} 
\cos(\theta) & -\sin(\theta)\\
\sin(\theta) & \cos(\theta)\\
\end{bmatrix}
\begin{bmatrix} 
x  \\
y \\
\end{bmatrix}
, & &
\theta \sim \text{Uniform}(0,2\pi). & &  \label{eq:rot} 
\end{align}
\item Sample control points for the spline in the new coordinate system via following sampling scheme:
\begin{enumerate}
    \item Sample the order of the  B\'ezier curve, $n$, from values $\{1,2,3\}$ with equal probability.
    \item Generate control points  $\mathbf{P}$=$\{(x^c_{0},y^c_{0})$,\ldots, $(x^c_{n},y^c_{n})\}$ such that:
\begin{align*}
&x^c_0=a,   x^c_{n}=b,\\
&x^c_k=u_{(k)},~~\text{if } n>1, ~k=1,\ldots,n-1, \text{ where } u_{(k)} \text{ is the } k_{th} \text{ order  } \\\nonumber
&\text{statistic among } u_1,\ldots,u_{n-1}, \text{  and } u_k  \sim \text{ Uniform}(a, b) ,\\ 
&y^c_j  \sim \text{Uniform}(c,d) , \forall j \in \{0,n\}.
\end{align*}\vspace{-0.5em}
 \item The B\'ezier curve $g_{\theta,\mathbf{P}}(x')$ in the new coordinated system is then determined by the control points, according to~\cite{mortenson1999mathematics}. 
 \item Sample $t\sim \text{Uniform}(l_{1},l_{2})$, shift the B\'ezier curve along the y-axis in the new coordinate system by adding $t$ to it. Denote the new B\'ezier curve as $g_{\theta,\mathbf{P}}(x')+t$. Assume $l_1<l_2$, $l_1$, $l_2$ are chosen properly such that all the curves that intersect with the target could be covered.  
 \end{enumerate}
\item If the proposed curve intersects with the target, divide the segment into two new segments by the  B\'ezier curve  $g_{\theta,\mathbf{P}}(x')+t$. Otherwise, return to step 1. 
\item   Transform the new segments and the B\'ezier curve  $g_{\theta,\mathbf{P}}(x')+t$ back to the original coordinate system.
\end{enumerate}
Note that we assume that $a < b, c <d$. If we choose $a, b, c, d$ appropriately, the proposed curves would  likely to be uniformly distributed over the target.  The selection of parameters $a, b, c, d$ would impact the effectiveness of the proposed curves. As illustrate in Figure \ref{fig: abcd}, if parameters $a, b$ are two small or two large, no matter how we lift the curve up and down, the proposed curve will not intersect with the target; if $a$ is not small enough or $b$ is not large enough, after rotation, it is likely to result in an invalid intersect; if $c$ is too small, or $d$ is too large, the segment of the curves that intersects with target will approximately be straight, it will lose the flexibility of the curve; with proper selection of $a, b, c, d$, after any arbitrary rotation, by lifting the curve up and down, the proposed curve will split the target successfully once it intersects with the target without losing the flexibility. Moreover, to maintain a high acceptance rate of the proposed curves, one of  possible settings of $l_1$, $l_2$ could be determined by letting $\min\{g_{\theta,\mathbf{P}}(x')\}+t<\max\{y'\}, \max\{g_{\theta,\mathbf{P}}(x')\}+t> \min\{y'\}$, which guarantees a tighter bound of B\'ezier curves that are intersect with the target. 

\subsection{Translational and rotational invariance}
 \label{sec:pSMSP}
\begin{figure}
\centering
\begin{subfigure}{.32\textwidth}
  \centering
 \includegraphics[width=0.85\linewidth]{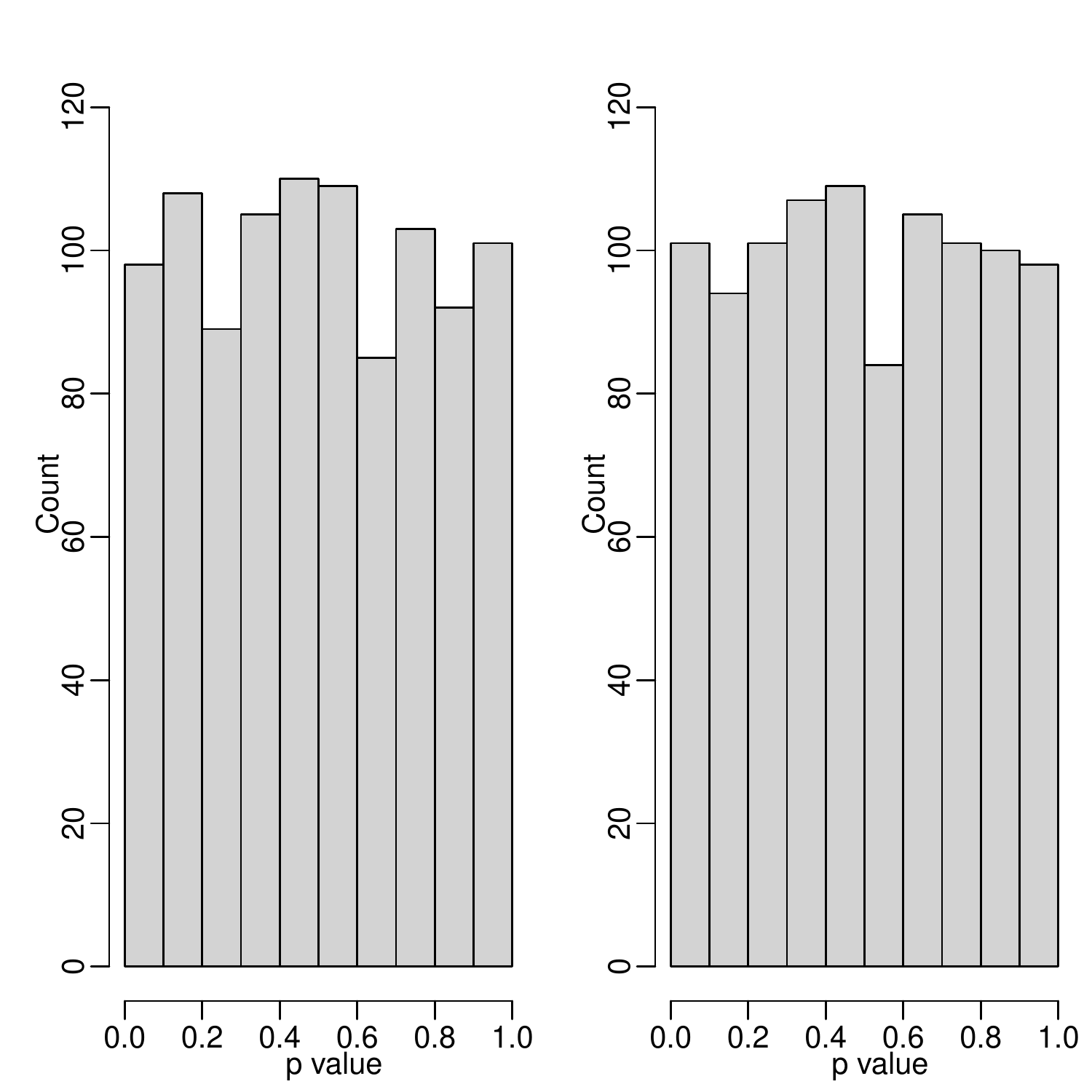}
  \caption{ }
  \label{cut1st2d:sub1}
\end{subfigure}
\begin{subfigure}{.65\textwidth}
  \centering
 \includegraphics[width=0.9\linewidth]{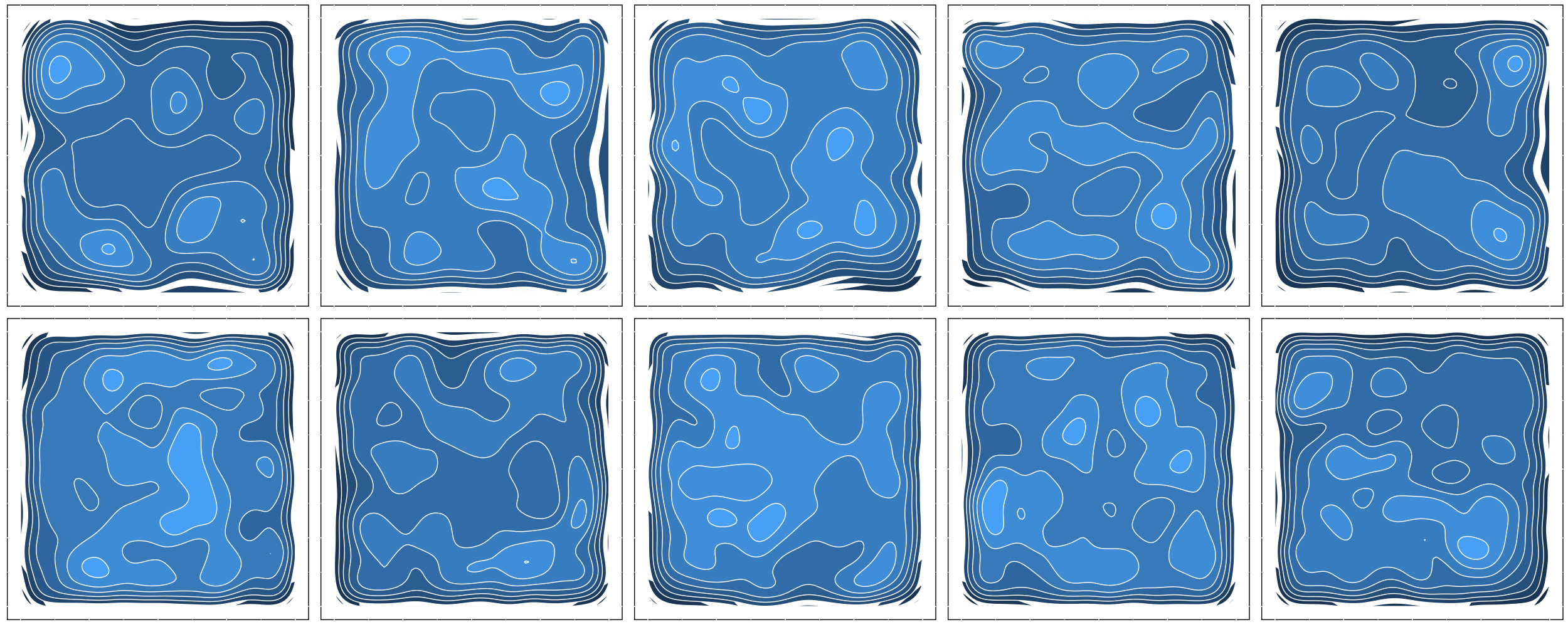}
  \caption{ }
  \label{cut1st2d:sub2}
\end{subfigure}
\caption{ \hspace{-0.15em} 
\textit{(a)} Frequency of $p$-values of bivariate uniformity tests on $1000$ experiments.  The \textit{left} panel displays the $p$-values of data  sampled from the first cuts, the \textit{right} panel presents the $p$-values of data sampled from a bivariate uniform distribution. \textit{(b)} The \textit{upper} panels are bivariate density plots of 5 randomly selected experiments, in which  data are sampled from the proposed first cuts on a unit square; the \textit{lower} panels show density of $5$ random draws of $5000$ points sampled from a uniform distribution on a unit square. 
}
\label{cut1st2d}
\end{figure}

Translational and rotational invariance are desirable properties for Bayesian nonparametric objects (as the inference becomes independent of the nature of the injection of the object into Euclidean space). For a bounded domain $W$, if $Y_t(W)$ is equal in distribution to $Y_t(TW)$ (where $T$ is an affine transformation), then translational and rotational invariance hold. We conduct a simulation study to explore invariance for the SMSP. The experiment is repeated $1000$ times with different random seeds. In each replicate, we  generate $5000$ random B\'ezier curves that intersect with the unit square $[0,1]^2$, with $a=-\sqrt{2}/2,
b=\sqrt{2}/2,c=-\sqrt{2}/2,d=\sqrt{2}/2$.  {\color{black} By selecting these parameters, the curves could intersect with the target cell after  arbitrary rotation, or after lifting the curves up and down within a reasonable interval.} We randomly sample one point from  each B\'ezier curve, the total number of sampled points is $5000$ in each replicate. Under the assumption of invariance, the sampled points lying inside the square are bivariate uniformly distributed. 
We then conduct a bivariate uniformity test  \citep{yang2017multivariate} on the sampled points for each experiment. The null hypothesis $\textit{H}_0$ is that the data are drawn from a bivariate uniform distribution on a square. The frequency of the $p$-values  is displayed in Figure~\ref{cut1st2d} (\emph{a}) \textit{left} panel, in which $950$ out of the $1000$ p-values are greater than $0.05$.  We also provide the p-values of the bivariate uniformity test on $1000$ random draws of $5000$ points sampled from a uniformly distribution on $[0,1]^2$ as a benchmark in Figure~\ref{cut1st2d} (\emph{a}) \textit{right} panel, and {\color{black} $95.1\%$} of the $p$-values are greater than $0.05$.  In Figure~\ref{cut1st2d} (\emph{b}) \textit{upper} panels, we provide the bivariate density plots of $5$ randomly drawn data sets from the $1000$ replicates. The density plots show that the distribution of data points are not concentrated on a certain small area, but on many small areas or the distribution is relatively flat over the square, which has a similar pattern as bivariate uniform distributions (e.g. Figure~\ref{cut1st2d} (\emph{b}) \textit{lower} panels). This indicates the sampled points may follow a bivariate uniform distribution, and provides empirical evidence of invariance.

Translational and rotational invariance of the SMSP also follow from our construction of the spline measure $\Lambda$. In the construction described in Section~2.3, the splines are parameterized by $\theta$ (angle),  $n$ (the order of the B\'ezier curve), $\mathbf{P}$ (control points) and $t$ (offset). Let $W$ be the set of predictors. In the construction, $\theta$, ($n$, $P$) and $t$  are independent. The control points are described in a coordinate system wherein the $x$ coordinates are parallel to the $x$ axis (and then these control points are rotated and translated by $\theta$ and $t$). Thus, before transformation these control points are also independent from $\theta$ and $t$. We can therefore decompose $\Lambda$ into the product measure $\Lambda^* \times \mu$, where $\Lambda^*$ is a measure on $\theta, t$, and $\mu$ is a measure on $n, P$. For any affine transformation $T$, if the points of a spline $\theta, t, n, P$ are transformed by $T$, the resulting spline will be described by the points of the spline $T(\theta,t), n, \mathbf{P}$,  where $(n, \mathbf{P})$ determines the shape of the spline and $T(\theta,t)$ represents the transformation. Thus, if $\Lambda^*$ is transitionally and rotationally invariant ($\Lambda^*(\cdot) = \Lambda^*\circ T(\cdot)$), $\Lambda(\cdot) = \Lambda^*\times\mu(\cdot) = (\Lambda^*\circ T)\times\mu(\cdot)= \Lambda^*\times\mu(T\cdot) = \Lambda(T\cdot)$, and translational and rotational invariance of $\Lambda$ follows. By the construction in Section~2.3,   $\Lambda^*$ is transitionally and rotationally invariant and so the result follows.

\subsection{Sequential Monte Carlo for SMSP inference}
We use techniques from relational modeling for  categorical data in order to specify a likelihood model for the SMSP. This specifies the same likelihood that was used in the random tessellation process~\citep{ge2019random}. 
Let $\boldsymbol{Y}_t( \boldsymbol{W})$ be an SMSP on the domain $W=\text{hull}\{\boldsymbol{v}_1,\ldots,\boldsymbol{v}_n\}$. 
We use $J_t$ to denote the number of subsets in  $\boldsymbol{Y}_t( \boldsymbol{W})$. 
Given the hyperparameter $\boldsymbol{\alpha}$, the likelihood function for  $\boldsymbol{Z}=(z_i)_{1\le i\le n}$ conditioned on $\boldsymbol{Y}_t( \boldsymbol{W})$ and  $\boldsymbol{V}=(\boldsymbol{v}_i)_{1\le i\le n}$  is as follows:
\begin{align}
 P(\boldsymbol{Z}|\boldsymbol{Y}_t( \boldsymbol{W}),\boldsymbol{V}\!,\boldsymbol{\alpha}) =  \prod_{j=1}^{J_t} \frac{B(\boldsymbol{\alpha} + \boldsymbol{m}_{j})}{B(\boldsymbol{\alpha})}. 
\label{likelihoodf}
 \end{align}
Here $B(\cdot)$ is the multivariate beta function, $\boldsymbol{m}_{j} = (m_{jk})_{1\le k\le K}$ and $m_{jk}=\sum_{i:h(\boldsymbol{v}_i)=j}\delta(z_{i} = k)$, and $\delta(\cdot)$ is an indicator function with $\delta(z_{i} = k)=1$ if $z_{i} = k$ and $\delta(z_{i} = k)=0$ otherwise.

Our objective is to infer the normalized posterior distribution of $\boldsymbol{Y}_t( \boldsymbol{W})$, denoted $\pi(\boldsymbol{Y}_t( \boldsymbol{W}))$.  
We use $\pi_{0}(\boldsymbol{Y}_t( \boldsymbol{W}))$ to represent the prior distribution of $\boldsymbol{Y}_t( \boldsymbol{W})$, which is the generative process of SMSP. 
Let $P(\boldsymbol{Z}|\boldsymbol{Y}_t( \boldsymbol{W}),\boldsymbol{V}\!,\boldsymbol{\alpha})$ denote the likelihood of the data given the spline partition $\boldsymbol{Y}_t( \boldsymbol{W})$ and the hyperparameter $\boldsymbol{\alpha}$. By Bayes' rule, the posterior of the partition at time $t$ is  
 $$\pi(\boldsymbol{Y}_t( \boldsymbol{W})|\boldsymbol{V}\!,\boldsymbol{Z},\boldsymbol{\alpha}) = \pi_{0}(\boldsymbol{Y}_t( \boldsymbol{W}))P(\boldsymbol{Z}|\boldsymbol{Y}_t( \boldsymbol{W}),\boldsymbol{V}\!,\boldsymbol{\alpha})/P(\boldsymbol{Z}|\boldsymbol{V}\!,\boldsymbol{\alpha}).$$
Here $P(\boldsymbol{Z}|\boldsymbol{V}\!,\boldsymbol{\alpha})$ denotes the marginal likelihood given data $\boldsymbol{Z}$. This marginal likelihood is intractable. Hence, the posterior distribution  of $\boldsymbol{Y}_t( \boldsymbol{W})$ does not admit a closed form. 
We develop an SMC algorithm to infer $\pi(\boldsymbol{Y}_t(\boldsymbol{W}))$, by taking advantage of the hierarchical structure of the SMSP prior. 
Our algorithm iterates between  the following three steps until a total budget $\tau$ is reached: resampling particles, propagation of particles and weighting of particles. 
\begin{itemize}
\item A resampling step  is conducted to prune particles with small weights.
\item We propagate particles via the generative process described in Algorithm~\ref{constructionofSMSP}, which is based on the generative process for the RTP~\citep{ge2019random}. As in~\cite{ge2019random}, we consider an approximation of the measure $\Lambda$ by replacing $\Lambda([W'])$ by $r$, where $r$ is the radius of the smallest circle covering the subset $W$, which is an upper bound for $\Lambda([W'])$. Note that in line 5 of Algorithm~\ref{constructionofSMSP}, we omit a potential scaling factor on the radius $r$ (this can be amortized by rescaling the prespecified budget). Firstly, we sample a subset $S$ from the set $\boldsymbol{Y}_{\tau_{m},m}(\boldsymbol{W})$ probability proportional to $r_S$.
     Secondly, we sample a B\'ezier curve $b$ from $[S]$ according to  $\Lambda(\cdot \cap [S])/\Lambda([S])$ using \emph{Section}~\ref{sec:cutplane}.
%\hspace{-0.5em}\smash{$\left.\rule{0pt}{2.0\baselineskip}\right]\ $}\mbox{\textit{IN PARALLEL}}
  Finally, we set $\boldsymbol{Y}_{\tau_m,m}(\boldsymbol{W}) \leftarrow \left(\boldsymbol{Y}_{\tau_m,m}(\boldsymbol{W})/\{ S \}\right) \cup \{S \cap b^-, S \cap b^+\!\}$.
 \item  In the weighting step, the weight update function is the ratio of likelihoods, as we propagate particles from the prior distribution. 
  We compute the normalized weights of each particle according to 
  \[
  \varpi_m \propto  \varpi_{m}P(\boldsymbol{Z}|\boldsymbol{Y}_{\tau_m,m}(\boldsymbol{W}),\boldsymbol{V},\alpha)/P(\boldsymbol{Z}|\boldsymbol{Y}_{\tau_m,m}'(\boldsymbol{W}),\boldsymbol{V},\alpha). 
  \]
Here the likelihood $P$ is define according to~\refp{likelihoodf}.
\end{itemize}
We obtain a list of weighted particles to  approximate the SMSP posterior $\sum_{m=1}^M\!\varpi_m\delta_{\boldsymbol{Y}_{\tau, m}(\boldsymbol{W})}$ at time $\tau$. Here $m$ is the particle index, $M$ is the total number of particles and $\varpi_m$ are the particle weights.   

Two methods are used to decrease the cost and complexity for computation. Firstly, we approximate $\Lambda([S])$ with the measure $\Lambda([\cdot])$ applied to the smallest closed circle containing $S$ in the computation of the lifetimes of subsets. Secondly, we use a pausing condition so that no cutting splines are proposed for subsets in which the labels of all predictors are same. We refer readers to \cite{ge2019random} for more details of these two methods and the framework of the SMC algorithm. In addition, we parallel the propagation step and the weighting step in the SMC algorithm by allocating samples across multiple CPUs (cores) to decrease the computational time.

\section{Experiments} \label{sec:exp}

In {\it Section} \ref{sec:yin-yang}, we conduct a simulation study to evaluate the performance of our SMC approach with respect to different number of cores and particles.  We also compare our SMSP to RTP, and SVMs in this section. In {\it Section} \ref{sec:macrophage} we evaluate our method via an HIV-1-infected human macrophage image data sets, and compare our method with modern machine learning approaches. We also explore the value of outputting shapes by computing the perimeter of the cells. Throughout our experiments, we set the likelihood hyperparameters for the SMSP to $\alpha_{k} = n_{k}/1000$, $n_{k}=\sum_{i}\delta(z_{i} = k)$, and use $\delta(\cdot)$ as is an indicator function with $\delta(z_{i} = k)=1$ if $z_{i} = k$ and $\delta(z_{i} = k)=0$ otherwise.

 \subsection{Simulations on a yin-yang dataset}
 \label{sec:yin-yang}

\begin{figure}[ht]
\centering
\begin{subfigure}{.155\textwidth}
  \centering
  \includegraphics[width=.95\linewidth]{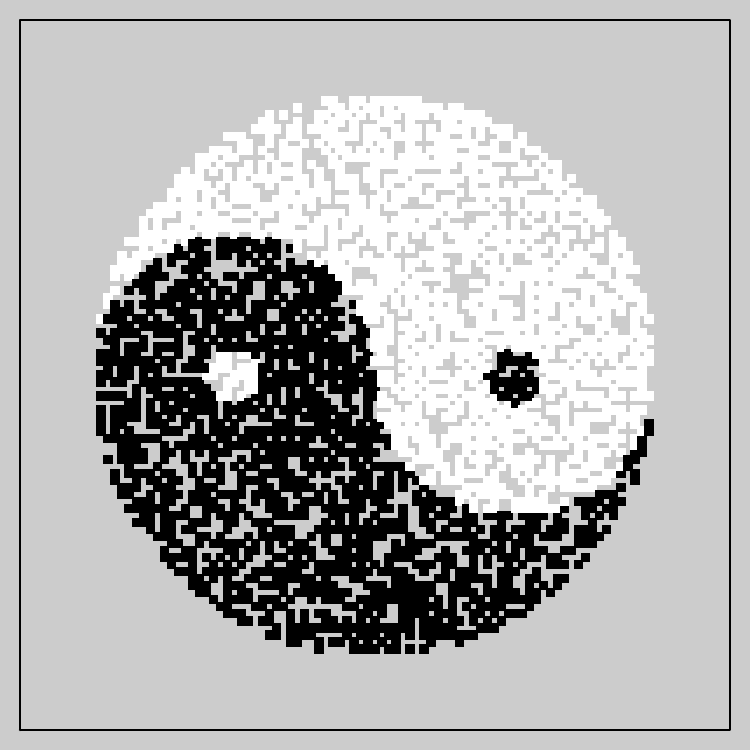}
  \caption{ }
  \label{fig:sub1}
\end{subfigure}
\begin{subfigure}{.82\textwidth}
  \centering
  \includegraphics[width=.95\linewidth]{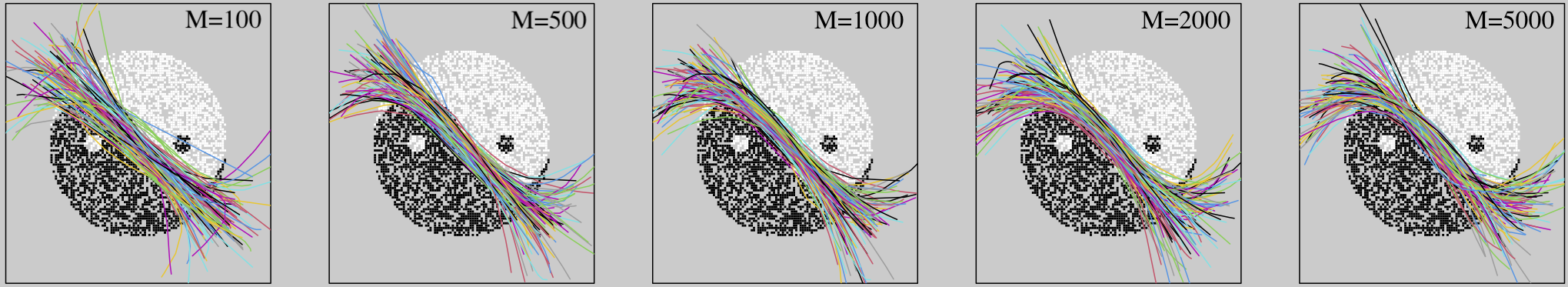}
  \caption{ }
  \label{fig:sub2}
\end{subfigure}
\caption{ \hspace{-0.15em} (\textit{a}) A view of the yin-yang training dataset, with black indicting label $1$ and white indicting label $2$. (\textit{b}) Best first cuts among different number of particles. Each experiment is repeated 100 times. 
}
\label{fig:test}
\end{figure}

\begin{figure}[h!]
\centering
\includegraphics[width=1.0\linewidth]{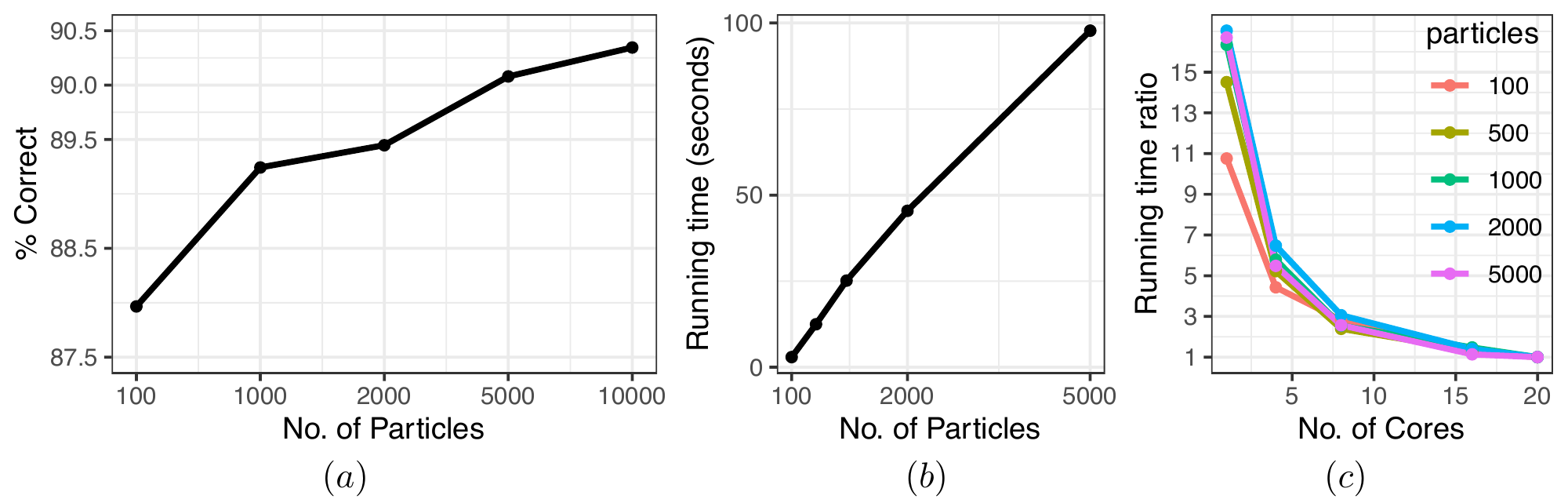}
   \caption{ (\textit{a}) \% correct of parallel SMC with different numbers of particles.  (\textit{b})  Running time ratio of SMC versus number of cores with different number of particles settings. The baseline is the running time with  $20$ cores.  (\textit{c}) Running time (seconds) of  SMC versus  different numbers of particles with  $16$ cores.  All experiments were run on Intel E5-2683 v4 Broadwell \@ 2.1Ghz machines. }
\label{fig:yinyang.sub3}
\end{figure}

We simulate a yin-yang dataset. First, we uniformly sample $10000$  points in the square $[-1,1]^2$, points falling outside of the circle $x^2+y^2=1$ are removed. Points satisfying $(x-0.5)^2+y^2<0.1^2$ or $x>0,y<0,(x-0.5)^2+y^2>0.25$ or $x<0,y<0,(x+0.5)^2+y^2>0.1^2$ or $x<0,y>0,(x+0.5)^2+y^2<0.25$ are labeled $1$ and the rest are labeled $2$. Figure~\ref{fig:test}(\textit{a}) provides a
visualization of the yin-yang training dataset, wherein points are colored black or white based on their label. 
We first conduct an experiment to evaluate the performance of SMC approach with respect to different number of particles. Each experiment is conducted with 4 cores and repeated $100$ times, and we allocate 60\% of the data items at random to a training set, the rest to the testing set. Figure~\ref{fig:test}(\textit{b}) displays the best first cuts with different number of particles. The first cut can generally capture the division of the yin-yang image when we use $N = 5000$ particles in SMC.
Figure \ref{fig:yinyang.sub3}{(\textit{a})} shows that the percentage correct of SMC increases with a larger value of particles, which indicates that a larger number of particles can improve the performance of the SMC algorithm. We note that due to the recursive nature of tree-based algorithms such as the SMSP, successful first cuts suggest good general performance of the algorithm. 

We conduct another experiment to evaluate the computational cost of our SMC approach. 
The experiments are run on Intel E5-2683 v4 Broadwell \@ 2.1Ghz machines. 
As indicated in Figure~\ref{fig:yinyang.sub3}(\textit{b}), the running time of SMC is a linear function of the number of particles. A larger number of particles increase the computational cost. But we can parallelize the propagation and weighting steps to decrease the time by allocating samples across different cores. The computational time  decreases more than $10$ times if we use $20$ cores (Figure~\ref{fig:yinyang.sub3}(\textit{c})). This shows one advantage of using our SMC approach over traditional Bayesian inference methods such as Markov chain Monte Carlo.

We conduct another set of experiments to compare our proposed method (SMSP) to RTP, and SVMs. We use $N=5000$ particles for both the SMSP and the RTP, and both methods are implemented with one cut. We implement two versions of SMSP, one by choosing the order of the B\'ezier curve $n$ from $1,2,3$ uniformly, and the other by fixing $n=3$. All methods are applied to 100 random train/test splits on the yin-yang dataset. We run sign test to compare the performance of  different methods.  Table \ref{table_yinyang} shows \% correct 
of all methods. 
It indicates that the SMSP with cubic proposal curves achieves better performance, compared with that proposing uniformly from straight, quadratic, cubic curves. Both SMSP methods perform significantly better than the RTP, and SVMs with linear, polynomial and sigmoid kernels. SVM with a radial kernel performs better than the SMSP. Note that for SMSP we only use a single cut, the performance would increase if we implement multiple cuts.

\begin{table}[ht]
\centering
\color{black}{
\caption{\color{black}{\% correct of SMSP with $n=3$ uniformly sampled from $1,2,3$ (SMSP), SMSP with $n=3$ (SMSP - $n=3$), RTP (RTP), and SVM with linear, radial, polynomial, and sigmoid kernels. Subscripts 1,2,3 indicate that the corresponding methods are the top 3 methods selected by the sign test. }}
\label{table_yinyang}
%\small
\begin{tabular}{l|rrrrrrrrrrrr}
  \hline
$\%$ correct & SMSP & SMSP  & RTP & SVM  & SVM  & SVM & SVM \\
  &   &   $n=3$ &  & linear & radial & polynomial & sigmoid \\
      \hline
  mean & $0.8951^3$  & $0.9019^2$  & 0.8859 &  0.8872 & $0.9667^1$  &    0.7513  &   0.8705 \\ 
  sd &  0.0091 &  0.0101 &  0.006 & 0.0046  &    0.0042 & 0.0067   & 0.0064   \\
   \hline
\end{tabular}
}
\end{table}

 \subsection{Experiment on an HIV-1–infected human macrophage}
 \label{sec:macrophage}

We evaluate the performance of the SMSP on a series of images of Human Immunodeficiency Virus Type I (HIV-1) infected human macrophage. We also display the results provided by some popular image segmentation approaches, such as  the Simple Linear Iterative Clustering (SLIC) Superpixels ~\citep{achanta2010slic, achanta2012slic}, zero version of the SLIC (SLICO) ~\citep{ren2003learning, lucchi2010fully}, affinity propagation clustering based on the SLIC superpixels (SLICAP) ~\citep{frey2007clustering,zhou2015image} and supervised classification algorithms methods, such as the K-nearest neighbor classifier (KNN), decision trees (DT), random forest (RF) (with $100$ trees), support vector machines (SVM). 

%{\color{black} Macrophages play a crucial role in innate and adaptative immunity in response to microorganisms and are an important cellular target during HIV-1 infection. They are the first line of defence of
%the organism against pathogens and, in response to the microenvironment, become differentially activated. The images are extracted from a real-time video showing HIV-1–infected human macrophage sensing its environment \citep{Gaudin}.  Images were collected on a spinning disk confocal microscope using a 100X objective.}

\begin{figure}[ht!]
\begin{subfigure}{.98\textwidth}
  \centering
  \includegraphics[width=0.99\linewidth]{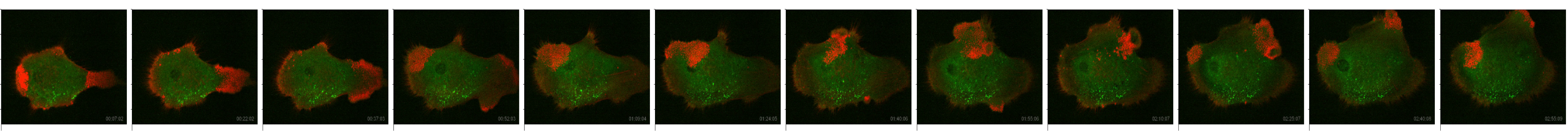}
  \caption{ }
  \label{fig:hiv.org}
\end{subfigure}
\begin{subfigure}{.98\textwidth}
  \centering
  \includegraphics[width=0.99\linewidth]{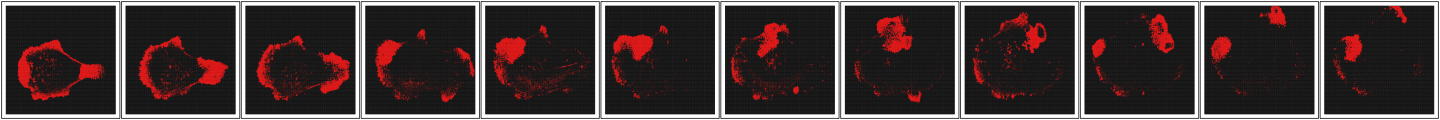}
  \caption{ }
  \label{fig:hivsub1}
\end{subfigure}
\begin{subfigure}{.98\textwidth}
  \centering
    \includegraphics[width=0.99\linewidth]{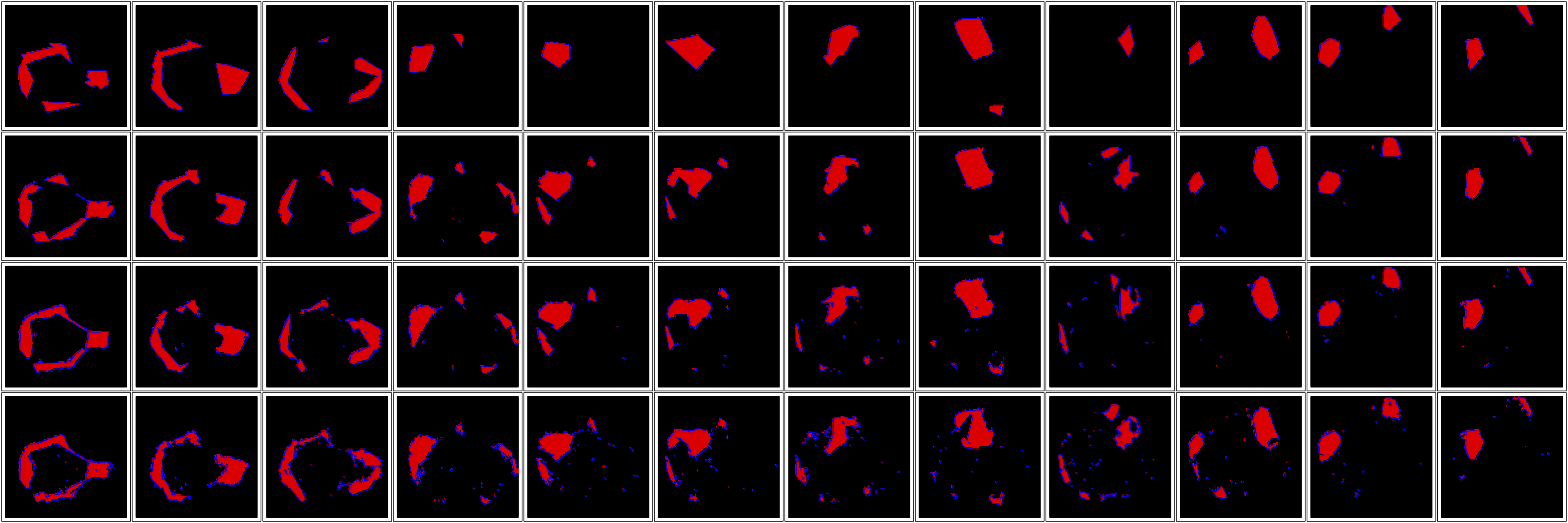}
  \caption{ }
  \label{fig:hiv_pmt}
\end{subfigure}
\caption{ \hspace{-0.15em} (\textit{a}) Original Images. (\textit{b}) Ground truth. (\textit{c}) SMSP produced image shapes (with shape boundaries in blue) with different budgets. Rows from top to bottom are the shapes produced by the SMSP with $\tau=10,~50,~100,~200$.}
\label{fig:hiv_org_pmt}
\end{figure}

{\color{black} Macrophages are a central aspect of the immune system, and are involved in the response to HIV-1 infection~\citep{Gaudin,Herbein:2010ug}. We consider data from an imaging experiment (confocal microscopy) in which HIV-1-infected macrophages interact with their environment~\citep{Gaudin}.} Images are taken from stills every second from the video, with $12$ images acquired (see Figure~\ref{fig:hiv_org_pmt}(\emph{a})). Each original image contains $600 \times 460$ pixels, and three colour channels. All images are converted to binary (black/white) image. For more detail on the imaging, see~\citeauthor{Gaudin}~\citeyear{Gaudin}. We reduce the size of each image to $20\%$ of its original size (\textit{i.e.} $120 \times 92$ pixels) and set the domain of the coordinates to $[-\frac{1}{2},\frac{1}{2}]\times [-\frac{1}{2},\frac{1}{2}]$, and  these images are assumed to be the ground truth (see Figure~\ref{fig:hiv_org_pmt}(\emph{b})).  Our goal is to capture the smooth shape of HIV-1–infected human macrophage for each image. In this experiment, we use $2000$ particles, $16$ cores, and set $\tau=\infty$ (\ie, running the SMSP until the pausing conditions are met for all partitions). The rest of the parameters are set to the default values described in the beginning of this section.

\section{Results} \label{sec:res} 
 
Macrophage infection is a major aspect of HIV-1 \citep{merrill1991hiv,cunningham1997hiv,hrecka2011vpx,koppensteiner2012macrophages}. Measures of HIV-1–infected human macrophages, such as the perimeters of infected cells may reveal aspects of HIV-1 etiology~\citep{Cas2017a}. To acquire these shapes within an image, we combine the curves that lie at the boundaries between subsets in the partition, and then mark the curve segments as interior boundaries if the $K$ nearest pixels on both sides of the curve have the same label, %if the $K$ nearest pixels on both sides of the curve have the same label,
and then remove the interior boundaries. This leaves only curves that delineate the boundary of the object. We sample points from the curves randomly and use the $K$-nearest neighbours algorithm, $K=10$, and restrict the distance within $\sqrt{\frac{1}{120^2}+\frac{1}{92^2}}$ to obtain the nearest pixels on both sides of the curve for each point. We also tried $K$-nearest neighbours without the distance restriction, but found that the boundaries were rougher.

\begin{figure}[ht!]
\centering
\begin{subfigure}{0.33\textwidth}
  \centering
   \includegraphics[width=1\linewidth]{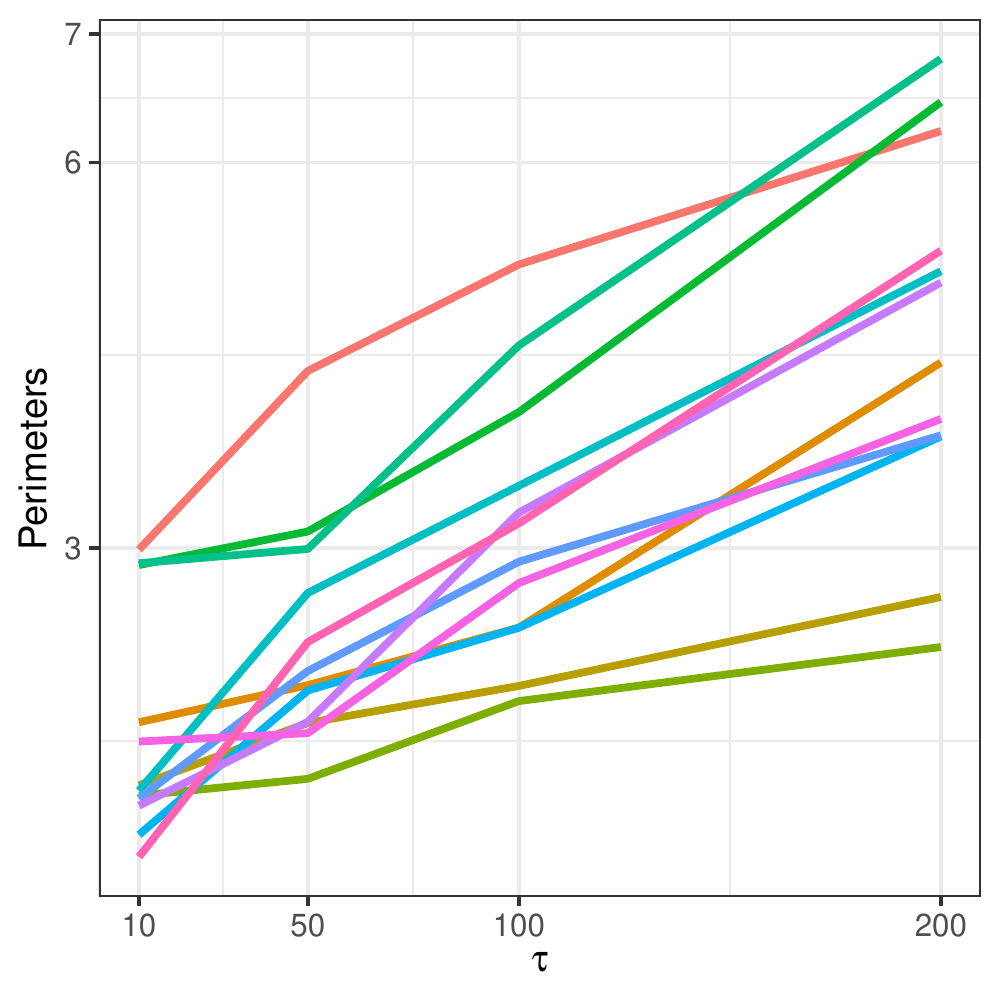}
  \caption{ }
  \label{fig:sub3}
\end{subfigure}%
\begin{subfigure}{0.66\textwidth}
  \centering
    \includegraphics[width=1\linewidth]{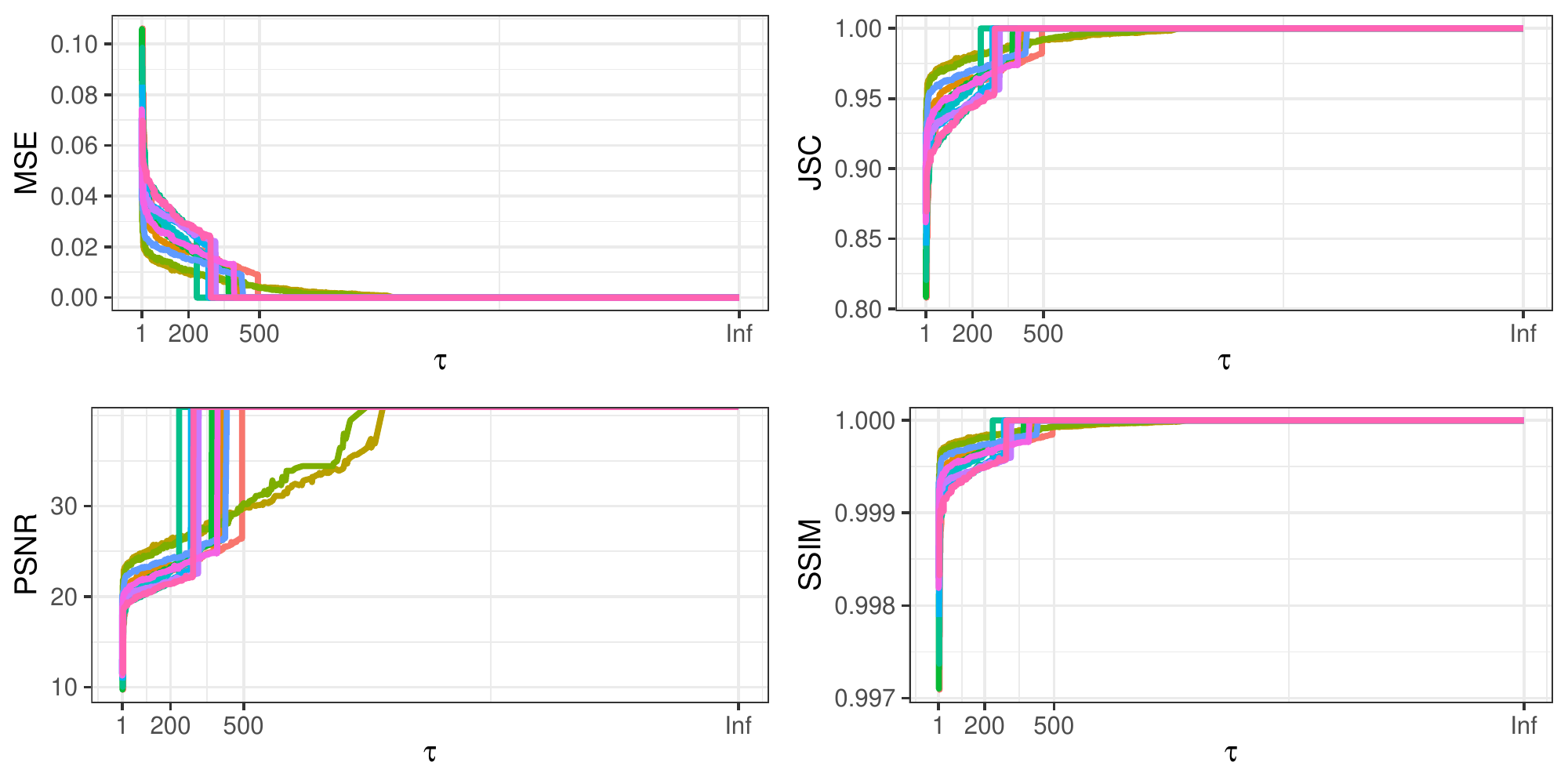}
  \caption{ }
  \label{hiv.summary}
\end{subfigure}%
\caption { (\textit{a}) Estimated perimeters of shapes as a function of  budget for images 1 to 12. Each line represents one image. (\textit{b}) Performance measures MSE, PSNR, JSC and SSIM as a function of  budget for hiv dataset. For MSE, lower is better and for PSNR, SSIM and JSC higher is better.}
\label{fig:hiv.summary}
\end{figure}

We then extract properties of the resulting shape. In this experiment we focus on the perimeter. Figure~\ref{fig:hiv_org_pmt}(\textit{c}) displays the shapes of the macrophage captured by the SMSP with different budgets, and Table~\ref{table_perimeter} displays the corresponding estimated perimeters. The 12 extracted images in the Table are labelled by \emph{I1}, \emph{I2}, $\ldots$. In Figure~\ref{fig:hiv_org_pmt}(\textit{c}), overplotting (in which multiple splines indicate the same border) arises on the boundaries. The degree of overplotting increases linearly with budgets: Figure~\ref{fig:hiv.summary}(\textit{a}). This suggests that the true perimeter of the shape is proportional to $\tilde p/\tau$, where $\tilde p$ is the perimeter of the shape found by the SMSP.

\begin{table}[ht]
\centering
\caption{Estimated perimeters of the shapes of the 12 images (I1 to I12) by the SMSP with different budgets. }
\label{table_perimeter}
%\small
\begin{tabular}{l|rrrrrrrrrrrr}
  \hline
$\tau$ & I1 & I2 & I3 & I4 & I5& I6 & I7 & I8 & I9 & I10 & I11 & I12\\
      \hline
10 & 2.99 & 2.87 & 2.88 & 1.11 & 0.77 & 1.05 & 1.00 & 1.50 & 0.60 & 1.65 & 1.16 & 1.08 \\ 
  50 & 4.38 & 3.13 & 2.99 & 2.65 & 1.89 & 2.04 & 1.65 & 1.56 & 2.27 & 1.93 & 1.64 & 1.21 \\ 
  100 & 5.21 & 4.06 & 4.58 & 3.48 & 2.38 & 2.89 & 3.28 & 2.73 & 3.19 & 2.38 & 1.93 & 1.81 \\ 
  200 & 6.24 & 6.47 & 6.81 & 5.15 & 3.87 & 3.88 & 5.07 & 4.01 & 5.31 & 4.44 & 2.62 & 2.23 \\
   \hline
\end{tabular}
\end{table}

\begin{table}[ht]
\centering
\caption{Mean of quantitative measures PSNR, MSE, JSC, SSIM, \% Correct on images over different methods.}
\label{table1}
\begin{tabular}{l|rrrrrrrrr}
  \hline
Methods & SVM & RF & DT & KNN & SLIC & SLICO & SLICAP & RTP &SMSP\\
  \hline
  PSNR &11.91 &17.83 &15.27 &15.10 &17.27 &17.27 &17.27 &Inf&Inf\\
 MSE &0.07  &0.02  &0.03  &0.03  &0.02  &0.02  &0.02  &0&0\\
 \% JSC & 87.51& 96.60 & 94.01 & 93.81  & 96.14  &96.15   &96.14 &1& 1\\
  \% SSIM&  99.83 & 99.97 & 99.95 & 99.94 & 99.96 & 99.96 & 99.96& 1&1 \\
 \% Correct & 93.30&  98.27 & 96.91 & 96.80 & 98.03 & 98.04 & 98.03 &100&100\\
   \hline
\end{tabular}
\end{table}

\begin{figure}[ht!]
\centering
\begin{subfigure}{0.50\textwidth}
  \centering
   \includegraphics[width=1\linewidth]{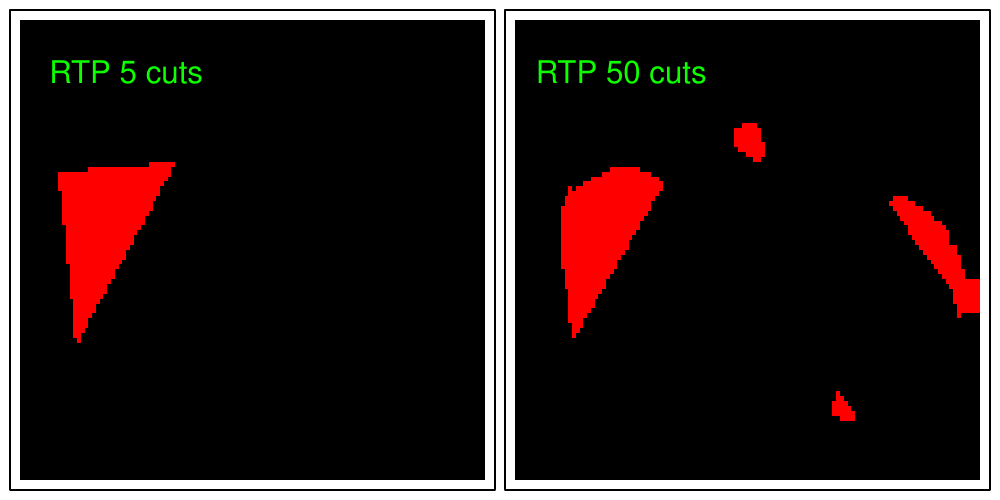}
  \label{fig:hiv_boundary1}
\end{subfigure}
\begin{subfigure}{0.50\textwidth}
  \centering
    \includegraphics[width=1\linewidth]{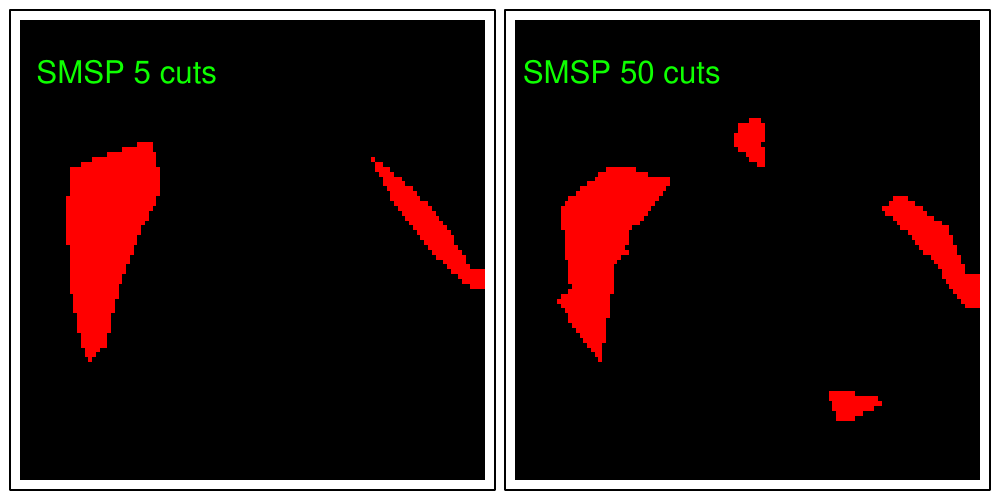}
  \label{hiv_boundary2}
\end{subfigure}
\caption{SMSP and RTP produced image shapes with different number of cuts on image I4. {\color{black}{The ground truth of this image is the fourth panel of Fig.~\ref{fig:hivsub1}.}}}
\label{fig:hiv_boundary}
\end{figure}

 We compare the ground truth and predicted images with different budgets via both visualization (see Supplementary Figure 1 in Appendix A of the \textit{Supplementary Material}) and quantitative measures (Figure~\ref{fig:hiv.summary}).
Figure~\ref{fig:hiv.summary}(\textit{b}) displays the mean squared error (\textit{MSE}), peak signal-to-noise ratio (\textit{PSNR}), Jaccard similarity coefficient (JSC) and structural similarity (SSIM) of SMSP as a function of budgets. Each line denotes a quantitative measure for one image {\color{black} (see Supplementary Appendix B for the formulas)}. The \textit{MSE}s decrease as the budget $\tau$ increase, and  becomes stable when $\tau$ exceed $20$. %number of cuts exceed $25$. %For some images, the \textit{MSE}s increase slightly with budget $\tau = \infty$, which indicates the shapes are overfitted in terms of \textit{MSE}. 
 The \textit{PSNR}s, JSCs and SSIM increase rapidly for  small budgets and  become stable when  $\tau>20$. %The shapes are slightly overfitted with budget $\tau = \infty$ in terms of \textit{PSNR}.
 Table \ref{table1} shows the average of different quantitative measures on the images across the methods. As indicated in Figure \ref{fig:hiv_boundary}, with a small number of cuts, the SMSP could capture the shape better than the RTP. With a large number of cuts, both the SMSP and RTP produce similar shapes, but SMSP could capture the subtle features better than the RTP.   Supplementary Figure 1(\textit{b}) displays the predicted images versus budgets, with ground truth (Supplementary Figure 1(\textit{a})), showing that a large budget can better capture the shape of the macrophage.
 Comparing Supplementary Figure~1(\textit{b}) to Supplementary Figure~1(\textit{c-d}), our proposed SMSP captures the shapes of the images comparable or  better to other methods.  Table \ref{table1} and Supplementary  Figure~1 indicate that our proposed SMSP with $\tau=\infty$ performs comparable to other methods in terms of the quantitative measures and visualization. The SVM in Table \ref{table1} uses radial basis kernels, the result of the SVM with other kernels in is given in Appendix A of the \emph{Supplementary Material}.

\section{Conclusion} \label{sec:con}
In this article, we present a novel Bayesian nonparametric method, the \emph{shape modeling with spline partitions}, for space partitioning. 
Our method extends the {random tessellation process}, by introducing a Bayesian nonparametric prior to partition two-dimensional Euclidean space with splines. 
We use B\'ezier curves to split space (\ie, approximating shapes), which enable us to represent more complex data-structure compared with existing methods~\citep{ge2019random, roy2008mondrian}. 
Our generative process of cutting splines provides evidence of  projectivity for the SMSP, which guarantees that our inference on model is well defined. 
To model the complex posterior distribution of SMSP, we develop a sequential Monte Carlo algorithm for inference. 
The proposed algorithm can avoid getting stuck around local modes of posterior and is easy to parallelize. We demonstrate empirical evidence of the consistency of our methods on simulated datasets. 
Our real data application shows that the SMSP has satisfactory performance in a task in which the smooth shape of HIV-1–infected human macrophage is captured, provided that enough computing resources are allocated (in terms of the number of particles and the computational budget $\tau$).
Our current work is limited to partition two-dimensional feature space, one possible future direction would be to extend the SMSP to capture shapes of multidimensional objects with cutting manifolds.

%\section*{Funding}
%SG was supported by the Shanghai Science and Technology Program (No. 21010502500), the startup fund of ShanghaiTech University, 
%SW was supported by the National Natural Science Funds of China (No. 12101333, No.62176068), the Natural Science Funds of Tianjin (No. 21JCQNJC00050),
%LTE was supported by a Natural Sciences and Engineering Research Council of Canada (NSERC) award DGECR/00118-2019 and a Michael Smith Health Research BC Scholar Award.
%
%
%
%\section*{Conflict of interest}
%The authors declare that they have no conflict of interest.

\bibliographystyle{spbasic}      % basic style, author-year citations
%\bibliography{document}

\begin{thebibliography}{40}
\providecommand{\natexlab}[1]{#1}
\providecommand{\url}[1]{{#1}}
\providecommand{\urlprefix}{URL }
\expandafter\ifx\csname urlstyle\endcsname\relax
  \providecommand{\doi}[1]{DOI~\discretionary{}{}{}#1}\else
  \providecommand{\doi}{DOI~\discretionary{}{}{}\begingroup
  \urlstyle{rm}\Url}\fi
\providecommand{\eprint}[2][]{\url{#2}}

\bibitem[{Achanta et~al.(2010)Achanta, Shaji, Smith, Lucchi, Fua, and
  S{\"u}sstrunk}]{achanta2010slic}
Achanta R, Shaji A, Smith K, Lucchi A, Fua P, S{\"u}sstrunk S (2010) Silc
  superpixels. Tech. rep.

\bibitem[{Achanta et~al.(2012)Achanta, Shaji, Smith, Lucchi, Fua, and
  S{\"u}sstrunk}]{achanta2012slic}
Achanta R, Shaji A, Smith K, Lucchi A, Fua P, S{\"u}sstrunk S (2012) Slic
  superpixels compared to state-of-the-art superpixel methods. IEEE
  Transactions on Pattern Analysis and Machine Intelligence 34(11)

\bibitem[{Berger(2012)}]{mjp}
Berger MA (2012) {An Introduction to Probability and Stochastic Processes}.
  Springer Texts in Statistics

\bibitem[{Bhattacharya and Dunson(2010)}]{bhattacharya2010nonparametric}
Bhattacharya A, Dunson DB (2010) Nonparametric {B}ayesian density estimation on
  manifolds with applications to planar shapes. Biometrika 97(4)

\bibitem[{Bu-Qing and Ding-Yuan(2014)}]{bu2014computational}
Bu-Qing S, Ding-Yuan L (2014) Computational Geometry: Curve and Surface
  Modeling. Elsevier

\bibitem[{Castellano et~al.(2017)Castellano, Prevedel, and Eugenin}]{Cas2017a}
Castellano P, Prevedel L, Eugenin AA (2017) {HIV}-infected macrophages and
  microglia that survive acute infection become viral reservoirs by a mechanism
  involving {B}im. Scientific Reports 7

\bibitem[{Chiu et~al.(2013)Chiu, Stoyan, Kendall, and Mecke}]{stoch}
Chiu SN, Stoyan D, Kendall WS, Mecke J (2013) {Stochastic Geometry and its
  Applications}. Wiley Series in Probability and Statistics

\bibitem[{Cunningham et~al.(1997)Cunningham, Naif, Saksena, Lynch, Chang, Li,
  Jozwiak, Alali, Wang, and Fear}]{cunningham1997hiv}
Cunningham AL, Naif H, Saksena N, Lynch G, Chang J, Li S, Jozwiak R, Alali M,
  Wang B, Fear W (1997) {HIV} infection of macrophages and pathogenesis of aids
  dementia complex: interaction of the host cell and viral genotype. Journal of
  Leukocyte Biology 62(1)

\bibitem[{Doucet et~al.(2000)Doucet, Godsill, and
  Andrieu}]{doucet2000sequential}
Doucet A, Godsill S, Andrieu C (2000) On sequential {M}onte {C}arlo sampling
  methods for {B}ayesian filtering. Statistics and Computing 10(3)

\bibitem[{Fan et~al.(2016)Fan, Li, Wang, Wang, and Chen}]{fan2016ostomachion}
Fan X, Li B, Wang Y, Wang Y, Chen F (2016) The {O}stomachion process. In:
  Proceedings of the Thirtieth Conference of the Association for the
  Advancement of Artificial Intelligence

\bibitem[{Fan et~al.(2018{\natexlab{a}})Fan, Li, and Sisson}]{fan2018binary}
Fan X, Li B, Sisson S (2018{\natexlab{a}}) The binary space partitioning-tree
  process. In: Proceedings of the 35th International Conference on Artificial
  Intelligence and Statistics

\bibitem[{Fan et~al.(2018{\natexlab{b}})Fan, Li, and
  Sisson}]{fan2018rectangular}
Fan X, Li B, Sisson S (2018{\natexlab{b}}) Rectangular bounding process. In:
  Proceedings of the 32nd Conference on Neural Information Processing Systems

\bibitem[{Fan et~al.(2019)Fan, Li, and Sisson}]{bspf}
Fan X, Li B, Sisson S (2019) Binary space partitioning forests. arXiv preprint
  190309348

\bibitem[{Fan et~al.(2020)Fan, Li, and SIsson}]{fan2020online}
Fan X, Li B, SIsson S (2020) Online binary space partitioning forests. In:
  International Conference on Artificial Intelligence and Statistics, PMLR, pp
  527--537

\bibitem[{Frey and Dueck(2007)}]{frey2007clustering}
Frey BJ, Dueck D (2007) Clustering by passing messages between data points.
  Science 315(5814)

\bibitem[{Gaudin(2011)}]{Gaudin}
Gaudin R (2011) Olympus BioScapes Digital Imaging Competition CIL:41568, Homo
  Sapiens, macrophage. CIL. Dataset.
  \url{https://doi.org/doi:10.7295/W9CIL41568}

\bibitem[{Gazit et~al.(1997)Gazit, Baish, Safabakhsh, Leunig, Baxter, and
  Jain}]{Gaz1997a}
Gazit Y, Baish JW, Safabakhsh N, Leunig M, Baxter LT, Jain RK (1997) Fractal
  characteristics of tumor vascular architecture during tumor growth and
  regression. Microcirculation 4(4)

\bibitem[{Ge et~al.(2019)Ge, Wang, Teh, Wang, and Elliott}]{ge2019random}
Ge S, Wang S, Teh YW, Wang L, Elliott LT (2019) Random tessellation forests.
  In: Advances in Neural Information Processing Systems

\bibitem[{George(1987)}]{george1987sampling}
George EI (1987) Sampling random polygons. Journal of Applied Probability 24(3)

\bibitem[{Gu et~al.(2012)Gu, Pati, and Dunson}]{gu2012bayesian}
Gu K, Pati D, Dunson DB (2012) Bayesian hierarchical modeling of simply
  connected 2d shapes. arXiv preprint arXiv:12011658

\bibitem[{Hannah and Dunson(2013)}]{hannah2013multivariate}
Hannah LA, Dunson DB (2013) Multivariate convex regression with adaptive
  partitioning. The Journal of Machine Learning Research 14(1)

\bibitem[{Herbein and Varin(2010)}]{Herbein:2010ug}
Herbein G, Varin A (2010) The macrophage in hiv-1 infection: from activation to
  deactivation? Retrovirology 7:33, \doi{10.1186/1742-4690-7-33}

\bibitem[{Hrecka et~al.(2011)Hrecka, Hao, Gierszewska, Swanson,
  {Kesik-Brodacka}, Srivastava, Florens, Washburn, and
  Skowronski}]{hrecka2011vpx}
Hrecka K, Hao C, Gierszewska M, Swanson SK, {Kesik-Brodacka} M, Srivastava S,
  Florens L, Washburn MP, Skowronski J (2011) Vpx relieves inhibition of
  {HIV}-1 infection of macrophages mediated by the {SAMHD1} protein. Nature
  474(7353)

\bibitem[{Kemp et~al.(2006)Kemp, Tenenbaum, Griffiths, Yamada, and
  Ueda}]{kemp2006learning}
Kemp C, Tenenbaum JB, Griffiths TL, Yamada T, Ueda N (2006) Learning systems of
  concepts with an infinite relational model. In: Proceedings of the 20th
  Conference on the Association for the Advancement of Artificial Intelligence

\bibitem[{Koppensteiner et~al.(2012)Koppensteiner, Brack-Werner, and
  Schindler}]{koppensteiner2012macrophages}
Koppensteiner H, Brack-Werner R, Schindler M (2012) Macrophages and their
  relevance in human immunodeficiency virus type i infection. Retrovirology
  9(1)

\bibitem[{Kurtek et~al.(2012)Kurtek, Srivastava, Klassen, and
  Ding}]{kurtek2012statistical}
Kurtek S, Srivastava A, Klassen E, Ding Z (2012) Statistical modeling of curves
  using shapes and related features. Journal of the American Statistical
  Association 107(499)

\bibitem[{Lakshminarayanan et~al.(2014)Lakshminarayanan, Roy, and
  Teh}]{lakshminarayanan2014mondrian}
Lakshminarayanan B, Roy DM, Teh YW (2014) Mondrian forests: {E}fficient online
  random forests. In: Proceedings of the 28th Conference on Neural Information
  Processing Systems

\bibitem[{Lakshminarayanan et~al.(2015)Lakshminarayanan, Roy, and
  Teh}]{lakshminarayanan2015particle}
Lakshminarayanan B, Roy DM, Teh YW (2015) Particle {G}ibbs for {B}ayesian
  additive regression trees. In: Proceedings of the 18th International
  Conference on Artificial Intelligence and Statistics

\bibitem[{Lin et~al.(2019)Lin, Mu, Cheung, and Dunson}]{lin2019extrinsic}
Lin L, Mu N, Cheung P, Dunson DB (2019) Extrinsic {G}aussian processes for
  regression and classification on manifolds. Bayesian Analysis 14(3)

\bibitem[{Lucchi et~al.(2010)Lucchi, Smith, Achanta, Lepetit, and
  Fua}]{lucchi2010fully}
Lucchi A, Smith K, Achanta R, Lepetit V, Fua P (2010) A fully automated
  approach to segmentation of irregularly shaped cellular structures in {EM}
  images. In: Proceedings of the International Conference on Medical Image
  Computing and Computer-Assisted Intervention, Springer

\bibitem[{Merrill and Chen(1991)}]{merrill1991hiv}
Merrill JE, Chen IS (1991) {HIV}-1, macrophages, glial cells, and cytokines in
  {AIDS} nervous system disease. The FASEB Journal 5(10)

\bibitem[{Mortenson(1999)}]{mortenson1999mathematics}
Mortenson ME (1999) Mathematics for computer graphics applications. Industrial
  Press Inc.

\bibitem[{Muller(1997)}]{muller1997surface}
Muller H (1997) Surface reconstruction: {A}n introduction. In: Scientific
  Visualization Conference, IEEE

\bibitem[{Nagel and Weiss(2005)}]{nagel2005crack}
Nagel W, Weiss V (2005) Crack {STIT} tessellations: {C}haracterization of
  stationary random tessellations stable with respect to iteration. Advances in
  Applied Probability 37(4)

\bibitem[{Rainforth and Wood(2015)}]{rainforth2015canonical}
Rainforth T, Wood F (2015) Canonical correlation forests. arXiv preprint
  150705444

\bibitem[{Ren and Malik(2003)}]{ren2003learning}
Ren X, Malik J (2003) Learning a classification model for segmentation. In:
  Proceedings 9th IEEE International Conference on Computer Vision

\bibitem[{Roy and Teh(2008)}]{roy2008mondrian}
Roy DM, Teh YW (2008) The {M}ondrian process. In: Proceedings of the 22nd
  Conference on Neural Information Processing Systems

\bibitem[{Tomita et~al.(2015)Tomita, Browne, Shen, Patsolic, Yim, Priebe,
  Burns, Maggioni, and Vogelstein}]{tomita2015random}
Tomita TM, Browne J, Shen C, Patsolic JL, Yim J, Priebe CE, Burns R, Maggioni
  M, Vogelstein JT (2015) Random projection forests. arXiv preprint 150603410

\bibitem[{Yang and Modarres(2017)}]{yang2017multivariate}
Yang M, Modarres R (2017) Multivariate tests of uniformity. Statistical Papers
  58(3)

\bibitem[{Zhou(2015)}]{zhou2015image}
Zhou B (2015) Image segmentation using {SLIC} superpixels and affinity
  propagation clustering. International Journal of Science and Research 4(4)

\end{thebibliography}

\end{document}